\documentclass[11pt,a4paper]{article}
\usepackage[hyperref]{acl2018}

\usepackage{amsmath}
\usepackage{array}
\usepackage{caption}
\usepackage{changepage}
\usepackage{float}
\usepackage{graphicx}
\usepackage{latexsym}
\usepackage{multirow}
\usepackage{setspace}
\usepackage{soul}
\usepackage{subcaption}
\usepackage{times}
\usepackage{todonotes}
\usepackage{verbatim}

\usepackage{url}

\aclfinalcopy 

\title{Learning Semantic Textual Similarity from Conversations}

\author{
Yinfei Yang\textsuperscript{$a$},
Steve Yuan\textsuperscript{$c$},
Daniel Cer\textsuperscript{$a$},
Sheng-yi Kong\textsuperscript{$a$},
Noah Constant\textsuperscript{$a$},
Petr Pilar\textsuperscript{$c$}, \\
\rm\textbf{Heming Ge\textsuperscript{$a$},
Yun{-}Hsuan Sung\textsuperscript{$a$},
Brian Strope\textsuperscript{$a$},
Ray Kurzweil\textsuperscript{$a$}} \AND
  {\rm\textsuperscript{$a$}Google Research}\\Mountain View, CA, USA \And
  {\rm\textsuperscript{$b$}Google}\\Cambridge, MA, USA \And
  {\rm\textsuperscript{$c$}Google}\\Zurich, Switzerland
}

\date{}

\newcolumntype{L}[1]{>{\raggedright\let\newline\\\arraybackslash\hspace{0pt}}m{#1}}

\begin{document}
\maketitle
\begin{abstract}
We present a novel approach to learn representations for sentence-level semantic similarity using conversational data.
Our method trains an unsupervised model to predict conversational input-response pairs.
The resulting sentence embeddings perform well on the semantic textual similarity (STS) benchmark and SemEval 2017's Community Question Answering (CQA) question similarity subtask. Performance is further improved by introducing multitask training combining the conversational input-response prediction task and a natural language inference task.
Extensive experiments show the proposed model achieves the best performance among all neural models on the STS benchmark and is competitive with the state-of-the-art feature engineered and mixed systems in both tasks.
\end{abstract}

\section{Introduction}

\begin{figure}
    \centering
    \includegraphics[width=.46\textwidth]{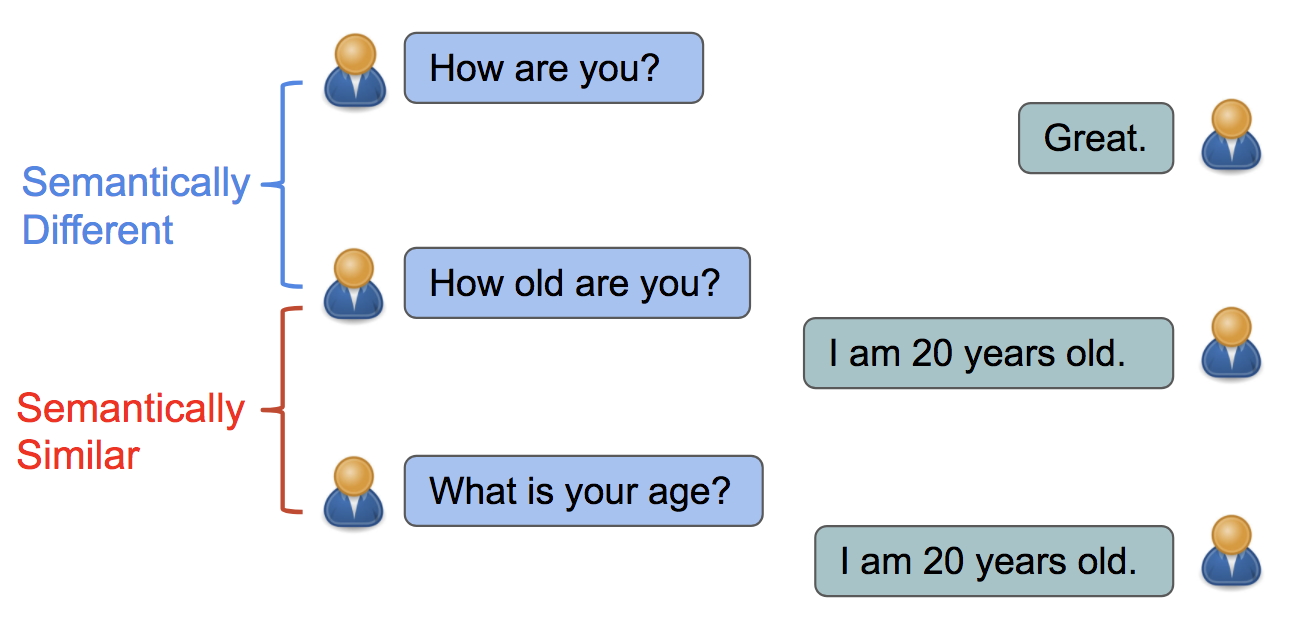}
    \caption{Sentences are semantically similar if they can be answered by the same responses. Otherwise, they are semantically different.}
    \label{fig:example}
\end{figure}

We propose a novel approach to sentence-level semantic similarity based on unsupervised learning from conversational data. We observe that semantically similar input sentences have a similar distribution of response sentences, and that a model trained to predict input-response relationships should implicitly learn useful semantic representations.
As illustrated in figure \ref{fig:example}, ``How old are you?''\ and ``What is your age?''\ are both questions about age, which can be answered by  similar responses such as ``I am 20 years old''.
In contrast, ``How are you?''\ and ``How old are you?''\ use similar words but have different meanings and lead to different responses.

Deep learning models have been shown to predict conversational input-response relationships with increasingly good accuracy \cite{efficientsmartreply,smartreply}. 
The internal representations of such models resolve the semantics necessary to predict the correct response across a broad selection of input messages. Meaning similarity between sentences then can be obtained by comparing the sentence-level representations learned by such models. We follow this approach, and assess the quality of the resulting similarity scores on the semantic textual similarity (STS) benchmark \cite{cer2017semeval} and a question-question similarity subtask from SemEval 2017's Community Question Answering (CQA) evaluation. The STS benchmark scores sentence pairs based on their degree of meaning similarity. The Community Question Answering (CQA) subtask B \cite{cqa} ranks questions based on their similarity with a target question.

We first assess representations learned from unsupervised conversational input-response pairs. We then explore augmenting our model with multi-task training over a combination of unsupervised conversational input-response prediction and supervised training on Natural Language Inference (NLI) data, which has been shown to yield useful general purpose representations \cite{infersent}. Unsupervised training over conversational data yields representations that perform well on STS and CQA question similarity. The addition of supervised SNLI data leads to further improvements and reaches state-of-the-art performance for neural STS models, surpassing training on NLI data alone.

\section{Approach}

This section describes the conversational learning task and our architecture for predicting conversational input-response pairs. We detail two encoding methods for converting sentences into sentence embeddings and describe multitask learning over conversational and NLI data.

\subsection{Conversational Learning Task using Input-Response Prediction}
\begin{figure}
    \centering
    \includegraphics[width=.46\textwidth]{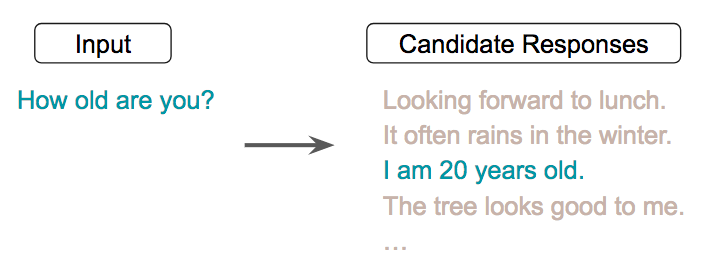}
    \caption{The response selection problem attempts to identify the correct response from a collection of candidate responses. We train using batch negatives with each candidate response serving as a positive example for one input and a negative sample for the remaining inputs.}
    \label{fig:response_prediction}
\end{figure}

We formulate the conversational learning task as response prediction given an input \cite{smartreply,efficientsmartreply}.
Following prior work, the prediction task is cast as a response selection problem.
As illustrated in figure \ref{fig:response_prediction}, the model $P(y|x)$ attempts to identify the correct response $y$ from $K-1$ randomly sampled alternatives.

\subsection{Model Architecture}

Our model architecture encodes input and response sentences into fixed-length vectors $u$ and $v$, respectively. Dot product is used to score the preference of an input described by $u$ for a response described by $v$. The dot product scores are converted into probabilities using a softmax function. Model parameters are trained to maximize the log-likelihood of the correct responses.

\begin{figure}
    \centering
    \includegraphics[width=.25\textwidth]{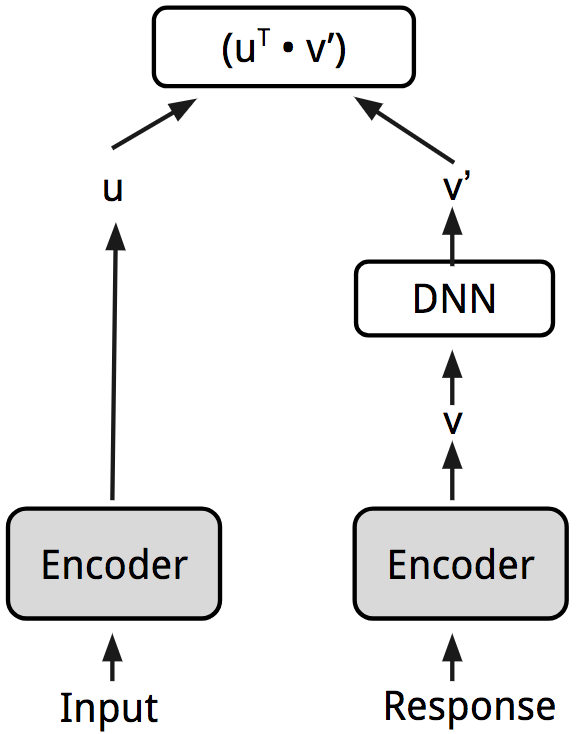}
    \caption{The basic input-response encoder model.  The sentence encoders in gray boxes use shared parameters. The DNN performs the mapping between the semantics of the input sentence and the expected response.}
    \label{fig:basic_model}
\end{figure}

Figure \ref{fig:basic_model} illustrates the input-response scoring model architecture. Tied parameters are used for the input and response encoders. In order to model the differences in meaning between inputs and responses, the response embeddings are passed through an additional feed-forward network to get the final response vector $v'$ before computing the dot product with the input embedding.\footnote{While feed-forward layers could have also been added to input encoder as well, early experiments suggested it was sufficient to add the additional layers to only one of the encoders.}

Training is performed using batches of K randomly shuffled input-response pairs. Within a batch, each response serves as the correct response to its corresponding input and the incorrect response to the remaining $K-1$ inputs in the batch. In the remaining sections, this architecture is referred to as the \emph{input-response model}.

\subsection{Encoders}

\begin{figure}[htb]
    \centering
    \begin{subfigure}[b]{0.18\textwidth}
        \includegraphics[width=\textwidth]{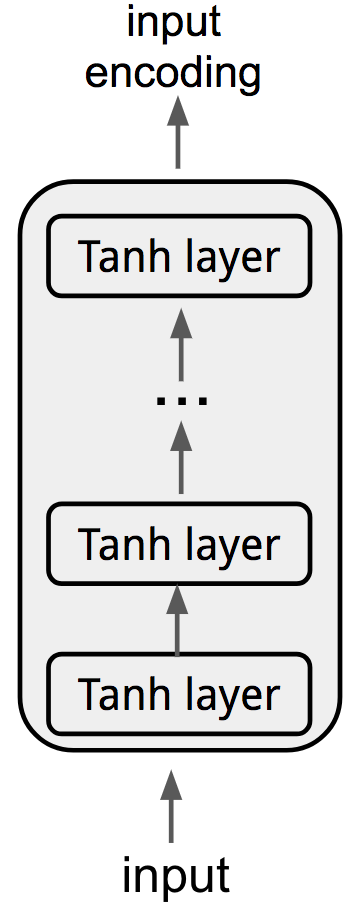}
        \caption{DAN encoder}
        \label{fig:dnn_encoder}
    \end{subfigure}
    \begin{subfigure}[b]{0.22\textwidth}
        \includegraphics[width=\textwidth]{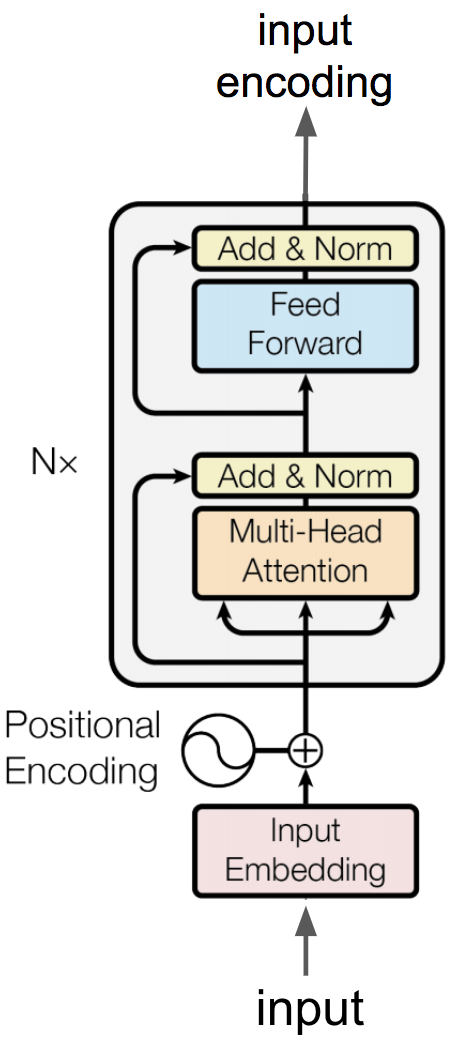}
        \caption{Transformer encoder}
        \label{fig:transformer_encoder}
    \end{subfigure}
    \caption{Model architectures for the DAN and Transformer sentence encoders.}
    \label{fig:encoders}
\end{figure}

Figure \ref{fig:encoders} illustrates the encoders we explore for obtaining sentence embeddings: DANs \cite{iyyer-EtAl:2015:ACL-IJCNLP} and Transformer \cite{transformer}\footnote{We tried other encoder architectures like LSTM and BiLSTM, but found they performed worse than transformer in our experiments.}.

\subsubsection{DAN}

The deep averaging network (DAN) computes sentence-level embeddings by first averaging word-level embeddings and then feeding the averaged representation to a deep neural network (DNN) \cite{iyyer-EtAl:2015:ACL-IJCNLP}. We provide our encoder with input  embeddings for both words and bigrams in the sentence being encoded. This simple architecture has been found to outperform LSTMs on email reply prediction \cite{efficientsmartreply}. The embeddings for words and bigrams are learned during the training of the input-response model. Our implementation sums the input embeddings and then divides by the $sqrt(n)$, where $n$ is the sentence length.\footnote{\texttt{sqrtn} is one of TensorFlow's built-in embedding combiners. The intuition behind dividing by $sqrt(n)$ is as follows: We want our input embeddings to be sensitive to length.
However, we also want to ensure that for short sequences the relative differences in the representations are not dominated by sentence length effects.} The resulting vector is passed as input to the DNN.

\subsubsection{Transformer}

Transformer \cite{transformer} is a recent network architecture that makes use of attention mechanisms to largely dispense with recurrence and convolutions. The approach achieves state-of-the-art performance on translation tasks and is available as open-source.\footnote{https://github.com/tensorflow/tensor2tensor}

While the original transformer contains an encoder and decoder, we only need the encoder component in our training procedure.
Each encoder layer consists of a multi-headed self-attention operation, followed by a feed-forward layer that processes each position independently (see figure \ref{fig:transformer_encoder}).
Positional information is captured by injecting a ``timing signal'' into the input embeddings based on sine/cosine functions at different frequencies.

As the transformer encoder output is a variable-length sequence, we reduce it to fixed length by averaging across all sequence positions.
Intuitively, this is similar to building a bag-of-words representation, except that the words have had a chance to interact with their contexts through the attention operations in the encoding layers.
In practice, we see that the learned attention masks focus largely on nearby words in the first layer, and attend to progressively more distant context in the higher layers.

\subsection{Multitask Encoder}

We anticipate that learning good semantic representations may benefit from the inclusion of multiple distinct tasks during training. Multiple tasks should improve the coverage of semantic phenomenon that are critical to one task but less essential to another. We explore multitask models that use a shared encoder for learning conversational input-response prediction and natural language inference (NLI). 
The NLI data are from the Stanford Natural Language Inference (SNLI) \cite{snli} corpus and are mostly non-conversational, providing a complementary learning signal.

\begin{figure}
    \centering
    \includegraphics[width=.45\textwidth]{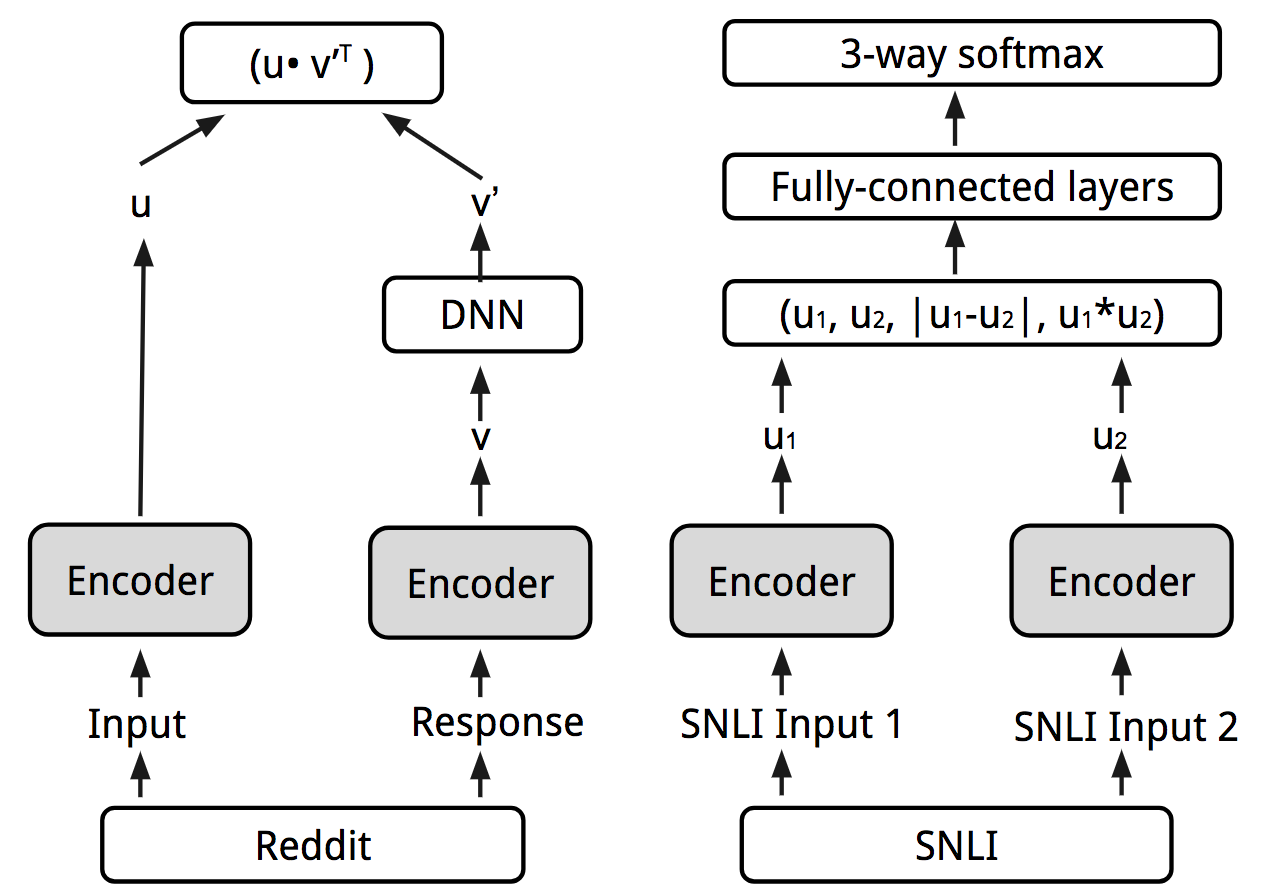}
    \caption{The architecture of the multitask model. The encoders in gray boxes use shared parameters.}
    \label{fig:snli_model}
\end{figure}

Figure \ref{fig:snli_model} shows the multitask model with SNLI\@.
We keep the input-response model the same, and build another two encoders for SNLI input pairs, sharing parameters with the input-response encoders.
Following \newcite{infersent}, we encode a sentence pair into vectors $u_1$, $u_2$ and construct a feature vector $(u_1, u_2, |u_1-u_2|, u_1*u_2)$.
The feature vector is fed into a 3-way classifier consisting of a feedforward neural network culminating in a softmax layer.
More specifically, we use a single layer network with 512 hidden units in all experiments, following previous work.

\section{Dataset}
\label{sec:dataset}

\begin{figure}
    \centering
    \includegraphics[width=.48\textwidth]{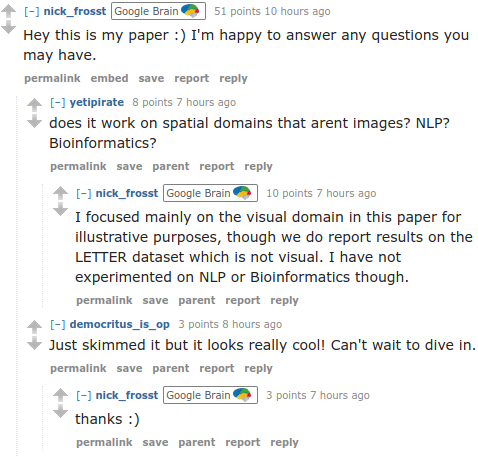}
    \caption{Example Reddit comment chain.}
    \label{fig:reddit_example}
\end{figure}

Our unsupervised model relies on structured conversational data.
The data for our experiments are drawn from Reddit conversations from 2007 to 2016, extracted by \newcite{al2016conversational}.
This corpus contains 133 million posts and a total of 2.4 billion comments, along with meta-data about the author of each comment and which comment it was replying to.
The comments are mostly conversational and well structured, which makes it a good resource for our model training. Figure \ref{fig:reddit_example} shows an example of a Reddit comment chain. 

Comment B is called a child of comment A if comment B replied to comment A.
We extract comments and their children to form the input-response pairs described above.
Several rules are applied to filter out the noisy data.
A comment is removed if any of the following conditions holds:
number of characters $\ge 350$, percentage of alphabetic characters $\le 70\%$, starts with ``https'', ``/r/'' or ``@'', author's name contains ``bot''. 
The total number of extracted pairs is around 600 million.

\section{Experiments}
We first evaluate the response prediction task with different encoders.
Then we discuss the results of supervised tasks with the model trained in the multitask framework.
Finally, we evaluate the learned encoders on the STS Benchmark \cite{cer2017semeval}.
We refer to the model trained over Reddit input-response pairs as \emph{Reddit model} and the multitask model as \emph{Reddit+SNLI}.

\subsection{Experiment Configuration}
Model configuration and hyperparameters were set based on prior experiments on Reddit input-response prediction and performance of the multi-task model on SNLI\@.
All input are tokenized and normalized before being fed into model.
For all experiments on the Reddit model and multitask model, we use SGD with batch size 128 and learning rate 0.01.
The total training steps is 40m steps for Reddit and 30m steps for multitask.
We tune the batch size to 256 and learning rate to 0.001 after 30m and 20m steps for Reddit and multitask respectively.
When training the multitask model, we initialize the shared parameters with a pretrained Reddit model.
We employ a distributed training system with multiple workers, where 95\% of workers are used to continue training the Reddit task and 5\% of workers are used to train the SNLI task. We use a sentence embedding size of 500 in all experiments, and normalize sentence embeddings prior to use in subsequent network layers.
The parameters were only lightly tuned to prevent overfitting on the SNLI task.

The configurations for encoders are taken from the default parameters from previous work.
For DAN, we employ a 3-layer DNN with layers of 300, 300, and 500 hidden units.
For the transformer encoder, our experiments make use of 6 attention layers, 8 attentions heads, hidden size 512, and filter size 2048.

\subsection{Response Prediction}

Following \newcite{efficientsmartreply}, we use precision at N (P@N) as an evaluation metric for the input-response prediction task.
Given an input, we select the true response (positive) and 99 random selected responses (negatives).
We score all 100 selected responses using the input-response scoring model and rank them accordingly.
The P@N score evaluates if the true response (positive) ranked in the top N responses.

Table \ref{tab:res_response_pred} shows the P@N results of Reddit models trained with different encoders, for N=1,3,10.
The DAN encoder (with n-grams) as used in \newcite{efficientsmartreply} can be considered a strong baseline. The transformer encoder outperforms DAN for all values of N\@. The transformer encoder achieves a P@1 metric of 65.7\% while the DAN encoder achieves only 56.1\%. Given its greater performance, we use a transformer encoder for the remainder of the experiments reported in this work.

\begin{table}
\centering
\caption{Precision at N (P@N) results of Reddit model with different encoders on the test set with 1 true response and 99 randomly selected negatives.}
    \label{tab:res_response_pred}
    \begin{tabular}{c | c c c} 
        \hline
        \textbf{}     &  P@1 & P@3 & P@10 \\ 
        \hline
        DAN           &  56.1 & 70.2 & 83.6\\
        Transformer   &  \textbf{65.7} & \textbf{78.7} & \textbf{89.8}\\
        \hline
    \end{tabular}
    
\end{table}

\subsection{SNLI}

Because the multitask model learns a shared encoder for the Reddit response prediction task and the SNLI classification task, we also report results on the SNLI task.
InferSent~\cite{infersent} is used as the baseline as it served as the inspiration for the inclusion of the SNLI task in the multitask model.
For comparison, we also list the results of Gumbel TreeLSTM~\cite{williams2017learning} which is the best sentence encoder based model,
and KIM Ensemble~\cite{chen2017nli} which is the current state-of-the-art. 
Sentence encoder based models first encode the two sentences in an SNLI input pair separately, and then feed the encodings into a classifier.
By comparison, other models also consider the interaction of the input pairs, for example using cross attention.
Our model falls under the category of sentence encoder based models.

\begin{table}
\centering
\caption{Performance on SNLI classification of the Reddit+SNLI model with the transformer encoder.}
    \label{tab:res_snli}
    \begin{tabular}{c | c} 
        \hline
        \textbf{}        &  Accuracy \\ 
        \hline
        Reddit+SNLI      &  84.1 \\
        InferSent        &  84.5 \\
        \hline
        \hline
        Gumbel TreeLSTM  &  86.0 \\
        KIM Ensemble     &  89.0 \\
        \hline
    \end{tabular}
    
\end{table}

Table \ref{tab:res_snli} shows the accuracy on the test set of the joint model and baselines.
The multitask model achieves 84.1\% accuracy and is close to the performance of InferSent. There are two significant differences between our model and prior work.
First, the proposed model learns all model parameters from scratch, including the word embeddings. 
InferSent uses large pre-trained word embeddings which could result in less out of vocabulary words.
In our work, the Reddit dataset is large enough that we do not need to use pre-trained word embeddings.\footnote{
Preliminary experiments with pre-trained embeddings on the Reddit dataset revealed no performance advantage over embeddings learned directly from the data.} 
Secondly, our multi-task model learns two tasks simultaneously, balancing performance between them, while InferSent only optimizes performance on SNLI.
As will be presented below, our multi-task model performs better on STS. We suspect the Reddit data acts as a regularizer to prevent the resulting sentence embedding from overfitting to the SNLI task, thus improving transfer performance on other tasks.

\subsection{STS Benchmark}

The proposed models encode text into a sentence-level embedding space. We evaluate the extent to which the embeddings encode sentence-level meaning using 
the Semantic Textual Similarity (STS) benchmark.
The benchmark includes English datasets from the SemEval/*SEM STS tasks
between 2012 and 2017. The data include 8,628 sentence
pairs from three categories: \textit{captions}, \textit{news}
and \textit{forums}. Each pair has a human-labeled degree of
meaning similarity, ranging from 0 to 5. The dataset is divided into
train (5,749), dev (1,500) and test (1,379).

\subsubsection{Evaluation}
We report results on two configurations for the Reddit
and multitask models. The first is ``out-of-the-box'' with no
adaptation for the STS task. In that case, we take the 
original sentence embedding vectors $u,v$ and score the
sentence pair similarity by the equation (\ref{eq:score})
together with an $\arccos$ that converts the cosine similarity
to distances that obey the triangle inequality.

\begin{equation}
  sim(u, v) =  - \arccos\bigg(\frac{u v}{||u|| ~ ||v||}\bigg)
\label{eq:score}
\end{equation}

For adaptation, the second configuration uses an additional
transformation matrix of the sentence embeddings. The matrix is parameterized using the STS training data. For both 
configurations, we also map the range of the scores to match
the range of the human labels using equation
(\ref{eqstsscore}).

\begin{equation}
  sim(u^*, v^*) = 5\Big(1 - \arccos\bigg(\frac{u^* v^*}{||u^*|| ~ ||v^*||}\bigg) / \pi\Big)
\label{eqstsscore}
\end{equation}

For model comparisons, we include the state-of-the-art neural
STS model CNN (HCTI)~\cite{shao2017hcti} and other systems in ~\newcite{cer2017semeval}: InferSent~\cite{infersent}, Sent2Vec~\cite{sent2vec}, SIF~\cite{sif}, PV-DBOW~\cite{pv_dbow}, C-PHRASE~\cite{c_phrase}, ECNU~\cite{ecnu} and BIT~\cite{bit}.
Note that the model trained on SNLI alone should be equivalent
to the InferSent baseline, with the exception that a
transformer encoder is used instead of a BiLSTM+MaxPooling.

Results on the STS Benchmark are listed in Table
\ref{tab:res_sts_reddit_snli}.
The columns show the accuracy of each model on the dev and
test sets. The un-tuned Reddit model is competitive and
outperforms many neural representation models. The encoder 
learned from Reddit conversations keeps text with
similar semantics close in embedding space. 

The ``out-of-the-box'' multitask model, Reddit+SNLI,  achieves
an $r$ of 0.814 on the dev set and 0.782 on test. Using a
transformation matrix to adapt the Reddit model trained without SNLI to STS, we achieve Pearson's $r$ of 0.809 on dev and 0.781 on
test. This surpasses the InferSent 
model, and is close to the performance of the best neural
representation approach, the CNN (HCTI) model.

The adapted multitask model achieves the best performance 
among all neural models, with an $r$ of 0.835 on the dev data
and 0.808 on test. The results are competitive with the
state-of-the-art feature engineered and mixed systems. The
proposed model is simpler and requires less human effort.

\begin{table}
\centering
\caption{Pearson's $r$ of Reddit+SNLI model with the transformer encoder on the STS Benchmark.}
    \label{tab:res_sts_reddit_snli}
    \begin{tabular}{c | c c} 
        \hline
        \textbf{}           & dev & test \\ \hline
        Reddit+SNLI tuned   & 0.835 & 0.808 \\
        Reddit+SNLI         & 0.814 & 0.782 \\
        Reddit tuned        & 0.809 & 0.781 \\
        Reddit              & 0.762 & 0.731 \\ \hline
        \multicolumn{3}{c}{Neural representation models} \\ \hline
        CNN (HCTI)          & 0.834 & 0.784 \\
        InferSent           & 0.801 & 0.758 \\
        Sent2Vec            & 0.787 & 0.755 \\
        SIF                 & 0.801 & 0.720 \\
        PV-DBOW             & 0.722 & 0.649 \\
        C-PHRASE            & 0.743 & 0.639 \\ \hline
        \multicolumn{3}{c}{Feature engineered and mixed systems} \\ \hline
        ECNU                & 0.847 & 0.810 \\
        BIT                 & 0.829 & 0.809 \\ \hline
    \end{tabular}
    
\end{table}

\subsubsection{Analysis}

\begin{table*}
\centering
\caption{Pearson's $r$ of the proposed models on the STS Benchmark with a breakdown by category.}
    \label{tab:res_sts_breakdown}
    \begin{tabular}{c | c| c c c || c| c c c } 
    \hline
        & \multicolumn{4}{c||}{dev} & \multicolumn{4}{c}{test} \\
        \cline{2-9}
        \textbf{}           & all & captions & forums & news & all & captions & forums & news \\ \hline
        Reddit              & 0.762 & 0.815 & 0.751 & 0.632 & 0.731 & 0.816 & 0.759 & 0.578 \\
        Reddit+SNLI         & 0.814 & 0.885 & 0.756 & 0.646 & 0.782 & 0.891 & 0.764 & 0.585 \\
        \hline               
        Reddit tuned        & 0.809 & 0.843 & 0.754 & 0.721 & 0.781 & 0.843 & 0.762 & 0.668 \\
        Reddit+SNLI tuned   & 0.835 & 0.888 & 0.759 & 0.731 & 0.808 & 0.894 & 0.767 & 0.667 \\
        \hline
    \end{tabular}
    
\end{table*}

To better understand the differences between the proposed models, we compare the Reddit and Reddit+SNLI models in more  detail, by breaking down the evaluation of STS Benchmark into
different categories.
In the test set, there are 625, 500, and 254 sentences pairs for \textit{image-captions}, \textit{news} and \textit{forums}, respectively.
The results for the sub-groups are shown in table \ref{tab:res_sts_breakdown}.

For the \textit{captions} category, adding the SNLI data improves the baseline Reddit model by about 8\%.
Interestingly, even with the adaptation to in-domain STS data, mixing in SNLI data still helps, as the tuned Reddit+SNLI model is 5\% higher than the tuned Reddit model.
The SNLI data helped our models with captions, presumably because most of the SNLI sentences are image captions, while Reddit doesn't contain much caption-style data.
In contrast, the performance of the models are similar in the other two categories.

The encoder learned from Reddit performs well on
general textual similarity. Model tuning with STS training
data helps adapt the representations of the sentence embeddings to emphasize the distinctions relevant to the task.
Also, given the large improvement from tuning with news data,
further improvements on the STS Benchmark could likely be achieved with more news data training.

\begin{figure}
    \centering
    \includegraphics[width=.48\textwidth]{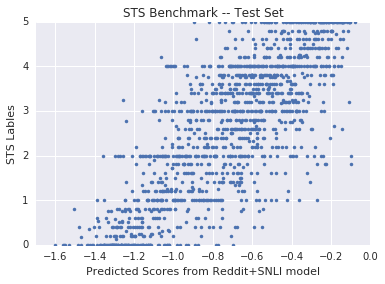}
    \caption{Plot of predicted scores vs.\ ground truth.}
    \label{fig:error_analysis}
\end{figure}

Figure \ref{fig:error_analysis} shows the plot of predicted 
similarity score versus the ground truth labels on the STS 
Benchmark test set.
The figure shows that the predicted scores are correlated 
with human judgment.
We list two good and two bad examples in table
\ref{tab:sts_example}.
With the good examples, the model has a relatively high score
for the semantically similar first pair, and a relatively low
score for the semantically different second pair.
In the first bad example, the model doesn't seem to weigh the
semantic distinction between boy and man as much as human
raters did. In the second bad example, apparently being about
whether to file Canadian tax returns was enough specificity 
for the model to consider the sentences similar. It's also possible to
read the second sentence as implying the first. As always,
more data in that domain would help the model make a better
choice.

\begin{table*}
\centering
\caption{Example model and human similarity scores on pairs from the STS Benchmark.}
    \label{tab:sts_example}
    \begin{tabular}{| c | c | c | L{10.5cm} |}
    \hline
         & Score & Label & \multicolumn{1}{c|}{STS Input Sentences} \\ \hline 
    Good & -0.51  & 4.2 & S1:  a small bird sitting on a branch in winter. \\
         &        &    & S2:  a small bird perched on an icy branch. \\ \hline
    Good & -1.23  & 0.0 & S1:  microwave would be your best bet. \\ 
         &        &    & S2:  your best bet is research. \\ \hline
    Bad  & -0.42  & 2.2 & S1:  a little boy is singing and playing a guitar.  \\
         &        &     & S2:  a man is singing and playing the guitar. \\ \hline
    Bad  & -0.45  & 1.0 & S1:  yes, you have to file a tax return in canada. \\
         &      &     & S2:  you are not required to file a tax return in canada if you have no \\
         &      &     &\hspace{0.6cm} taxable income. \\ \hline
    \end{tabular}
\end{table*}

\subsection{How Much Data for the Supervised Task?}

The experiments in the previous section show that supervised in-domain data can be used to improve performance on domain-specific tasks.
However, supervised data is difficult to obtain, especially on the order of SNLI's 570,000 sentence pairs.
In order to learn how much supervised data is needed,
we train multitask models with Reddit and varying amounts of SNLI data, ranging from 10\% to 90\% of the full dataset.

\begin{figure}
    \centering
    \includegraphics[width=.48\textwidth]{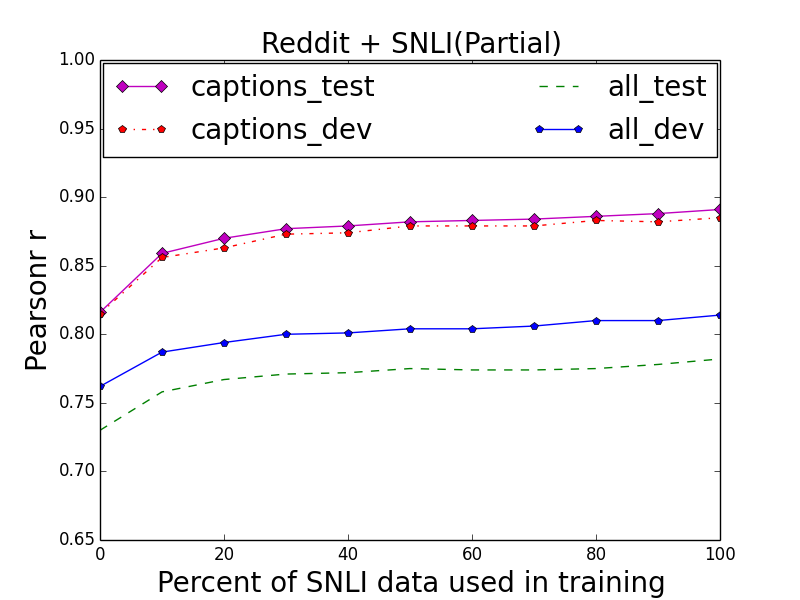}
    \caption{Pearson's $r$ on the STS Benchmark for the multitask model trained with Reddit and varying amounts of SNLI data.}
    \label{fig:reddit_percent_snli}
\end{figure}

Figure \ref{fig:reddit_percent_snli} shows the STS Benchmark results for all data and for captions data only, on both dev and test sets.
When first adding the SNLI data into the training task, Pearson's $r$ increases rapidly across all four tasks.
Even with only 10\% of the SNLI data, $r$ reaches around 0.85 for captions data in both of dev and test.
The curves mostly flatten out after using 40\% of the data, with performance only improving slightly with more data.
This indicates that the encoder learned from Reddit alone can be adapted to domain specific tasks efficiently using a small set of in-domain data.

\subsection{CQA Subtask B}

To further validate the effectiveness of sentence representations learned from conversational data, we assess the proposed models on subtask B of the SemEval Community Question Answering (CQA) task~\cite{cqa}.
In this task, given an ``original'' question $Q$, and the top ten related questions from a forum ($Q_1, \dots , Q_{10}$) as retrieved by a search engine, the goal is to rank the related questions according to their similarity with respect to the original question.
Mean average precision (MAP) is used to evaluate candidate models.

Each pairing of an original question and a related question ($Q, Q_i$) is labeled ``PerfectMatch'', ``Relevant'' or ``Irrelevant''.
Both of ``PerfectMatch'' and ``Relevant'' are considered as \textit{good} questions.
A candidate model should rank \textit{good} questions above the ``Irrelevant'' questions.
The task also allows models to use comments and user profiles as additional contextual features.

Similar to the STS experiments, we use cosine similarity between the original question and related question, without considering any interaction between the two questions, or any contextual features.
Given a related question $Q_i$ and its original question $Q$, we first encode them into vectors $u_i$ and $u$.
Then the related questions are ranked based on the cosine similarity $cos(u_i, u)$ with respect to the original question.
Results are shown in table \ref{tab:res_cqa_b}. 
SimBow~\cite{charlet2017simbow} and KeLP~\cite{filice2017kelp}, which are the best systems on the 2017 task, are used as baselines\footnote{In the competition, each team can submit one primary run and two contrastive runs. Only the primary run is used for the official ranking.}.
Without any tuning on the training data provided by the task, both models show competitive performance.
Reddit+SNLI model outperforms SimBow-primary, which is ranked first in the official ranking of the 2017 task.

\begin{table}
\centering
\caption{Mean Average Precision (MAP) of the proposed models on CQA subtask B.}
    \label{tab:res_cqa_b}
    \begin{tabular}{c | c} 
        \hline
        \textbf{}           & MAP  \\ \hline
        KeLP-contrastive1   & 49.00 \\
        SimBow-contrastive2 & 47.87 \\
        SimBow-primary      & 47.22 \\
        \hline
        Reddit              & 47.07  \\
        Reddit+SNLI         & 47.42 \\
        \hline
    \end{tabular}
    
\end{table}

\section{Related Work}


The STS task was first introduced by \newcite{agirre2012semeval}. 
Early methods focused on lexical semantics, surface form matching and basic syntactic similarity \cite{bar2012ukp,jimenez2012soft}.
More recently, deep learning based methods became competitive \cite{shao2017hcti,tai2015improved}. One approach to this task is to train a general purpose sentence encoder and then calculate the cosine similarity between the encoded vectors for the pair of sentences.
The encoding model can be directly trained on the STS task \cite{shao2017hcti} or it can be trained on an alternative supervised~\cite{infersent} or unsupervised~\cite{sent2vec} task that produces sentence-level embeddings.
The work described in our paper falls into the latter category, introducing a new unsupervised task based on conversational data that achieves good performance on predicting semantic similarity scores. Training on conversational data has been previously shown to be effective at email response prediction \cite{smartreply,efficientsmartreply}. We extend prior work by exploring the effectiveness of representations learned from conversational data to capture more general-purpose semantic information. The approach is similar to Skip-Thought vectors \cite{NIPS2015_5950}, which learn sentence-level representations through prior and next sentence prediction within a document, but with our prior and next sentences being pulled from turns in a conversation.

\section{Conclusion}
In this paper, we present a response prediction model which learns a sentence encoder from conversations.
We show that the encoder learned from input-response pairs performs well on sentence-level semantic textual similarity.
The basic conversation model learned from Reddit conversations is competitive with existing sentence-level encoders on public STS tasks.
A multitask model trained on Reddit and SNLI classification achieves the state-of-the-art for sentence encoding based models on the STS Benchmark task.
Finally, we show that even without any task-specific training, the Reddit and Reddit+SNLI models are already competitive on CQA subtask B.

\section*{Acknowledgments}

We thank our teammates from Descartes and other Google groups for their feedback and suggestions, particularly Dan Gillick and Raphael Hoffman.

\bibliography{acl2018}
\bibliographystyle{acl_natbib}

\end{document}